\documentclass[letterpaper, 10 pt, conference]{ieeeconf}  

\IEEEoverridecommandlockouts                              

\usepackage{amsmath,amssymb,amsfonts}
\usepackage{algorithmic}
\usepackage{graphicx}
\usepackage{textcomp}
\usepackage{xcolor}
\usepackage{makecell}
\usepackage{multirow} 
\usepackage{tabularx} 
\usepackage{svg}
\usepackage{hhline}
\let\labelindent\relax
\usepackage{enumitem}
\usepackage{siunitx}
\usepackage{booktabs} 
\usepackage{subcaption}
\usepackage{hyperref}
\usepackage{fancyhdr}
\usepackage{stfloats}

\title{Improving Out-of-Distribution Generalization of Trajectory Prediction \\for Autonomous Driving via Polynomial Representations
}

\author{Yue Yao$^{1,2}$, Shengchao Yan$^3$, Daniel Goehring$^2$, Wolfram Burgard$^4$, Joerg Reichardt$^1$
\thanks{The authors are with $^1$Continental AG, $^2$Freie Universität Berlin, $^3$Department of Computer Science, University of Freiburg, $^4$Department of Engineering, University of Technology Nuremberg.}
}

\begin{document}

\maketitle

\begin{abstract}
Robustness against Out-of-Distribution (OoD) samples is a key performance indicator of a trajectory prediction model. However, the development and ranking of state-of-the-art (SotA) models are driven by their In-Distribution (ID) performance on individual competition datasets.
We present an OoD testing protocol that homogenizes datasets and prediction tasks across two large-scale motion datasets. We introduce a novel prediction algorithm based on polynomial representations for agent trajectory and road geometry on both the input and output sides of the model. With a much smaller model size, training effort, and inference time, we reach near SotA performance for ID testing and significantly improve robustness in OoD testing. Within our OoD testing protocol, we further study two augmentation strategies of SotA models and their effects on model generalization.  Highlighting the contrast between ID  and OoD performance, we suggest adding OoD testing to the evaluation criteria of trajectory prediction models.
\end{abstract}

\section{Introduction}

Trajectory prediction is essential for autonomous driving, with robustness being a key factor for practical applications. The development of trajectory prediction models is catalyzed through public datasets and associated competitions, such as Argoverse 1 (A1) \cite{chang_argoverse_2019}, Argoverse 2  (A2) \cite{wilson_argoverse2_2021}, Waymo Motion (WO) \cite{ettinger_waymo_2021}. These competitions provide a set of
standardized metrics and test protocol that scores prediction
systems on test data withheld from all competitors. This is
intended to ensure two things: objective comparability of
results and generalization ability due to the held-out test set. 

Among the top-performing models based on deep learning, we observe a trend to ever more parameter-rich and expressive models \cite{zhou_hivt_2022, shi_mtr++_2023, zhou_query_2023} - trained specifically with the data of each competition. This begs the question of whether the stellar performance of these models is due to their ability to adapt to each dataset specifically, i.e., over-fit.
Prediction competitions try to guard against over-fitting by scoring models with test samples that are withheld from competitors. However, these test examples still share similarities with the training samples, such as the sensor setup, map representation, post-processing, geographic, and scenario selection biases employed in dataset creation. Consequently, the test scores reported in each competition are examples of \emph{In-Distribution} (ID) testing.

\begin{figure}[!h]
\centering
\includegraphics[width=3.2in]{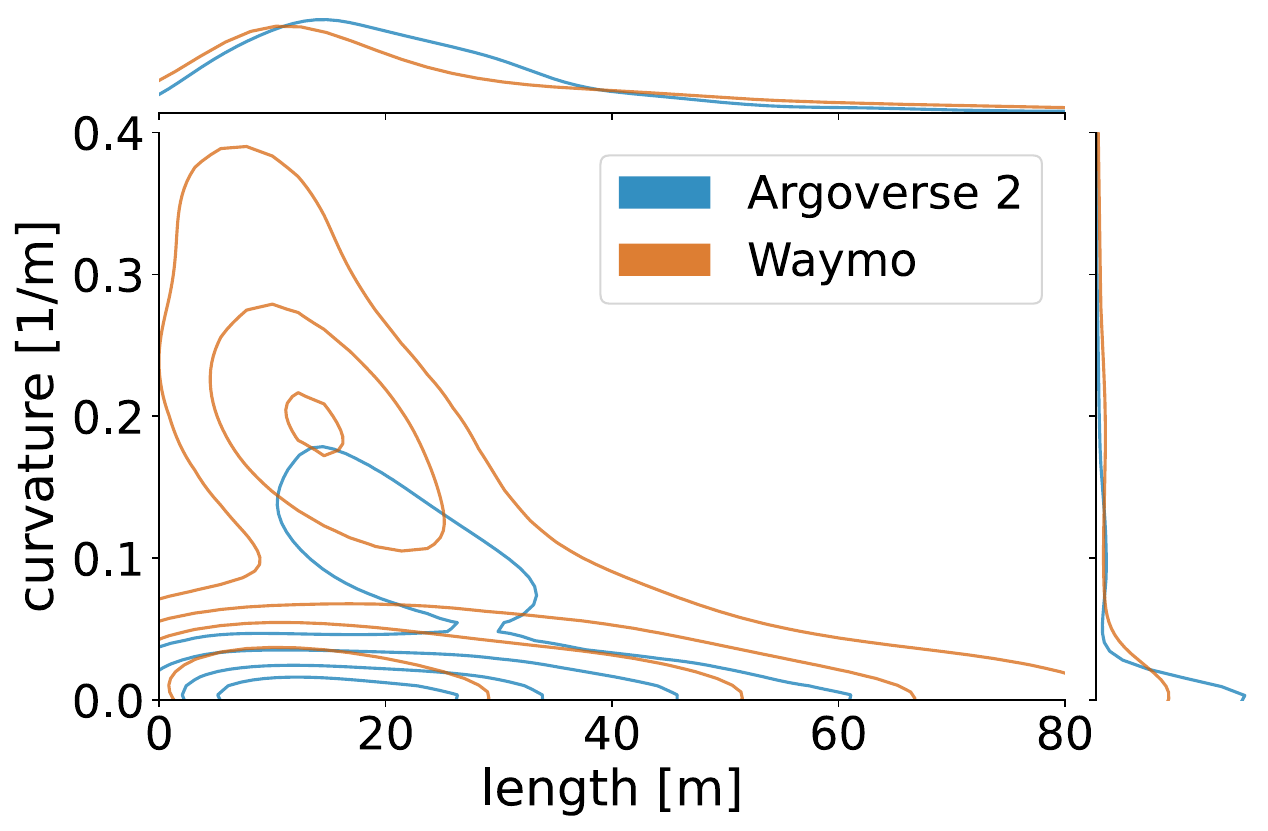}
\caption{Kernel density plot of the maximum absolute curvature and length for $5000$ random lane segments in A2 and WO. Contours indicate the $20$-th, $40$-th, $60$-th, and $80$-th percentiles, respectively.}
\label{fig:distribution_shift}
\vspace{-0.5em}
\end{figure}

For practical application, the performance of trajectory prediction should be independent of these shared biases in a single dataset and measured by \emph{Out-of-Distribution} (OoD) testing, e.g., across different datasets. As an example, Figure \ref{fig:distribution_shift} clearly visualizes the distribution shift caused by different map generation processes between A2 and WO. However, the efforts for cross-dataset evaluation are hampered by the in-homogeneous data format and task details among datasets and competitions.
Notable distinctions are the different lengths of scenario, observation history, and prediction horizon in each dataset.

Therefore, in this work, we attempt to homogenize the data formats and prediction tasks to enable OoD testing across two large-scale motion datasets: A2 and WO. Based on our OoD testing, we further explore the possibilities to enhance model robustness against OoD samples.

\begin{figure*}[!h]
\centering
\includegraphics[width=6.8in]{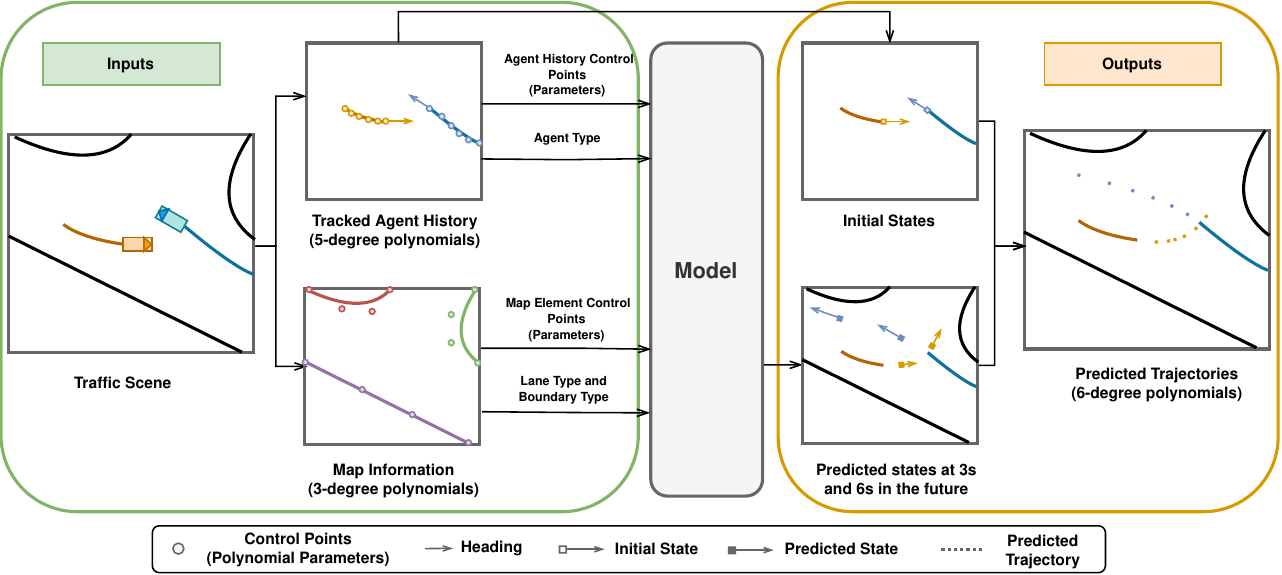}
\caption{Our proposed model architecture. \textbf{Inputs:} Agent histories and road geometry are both represented via polynomials. \textbf{Outputs:} The tracked initial states and predicted states are fused into one polynomial trajectory prediction, ensuring continuity of past observation and future prediction.}
\label{fig:pipeline}
\vspace{-1.0em}
\end{figure*}

One strategy to improve the robustness of any deep learning model is to increase the amount of training data. Bahari et al. \cite{bahari_vehicle_2022} have shown that generalization performance can be improved by augmenting training data with programmatically created variations. While such an approach does increase generalization ability, it further increases training effort and cannot guarantee generalization. Further, ensuring that a data augmentation strategy covers all possible variations is not easy.

We follow a complementary strategy and instead of providing more training data, limit the expressiveness of our model by constraining its input and output representation. This comes with the added benefit of a much smaller model size, reduced training effort and faster inference times. At the same time, we retain near state-of-the-art (SotA) prediction performance for ID test cases and show significantly improved robustness in OoD testing. Our key contributions are as follows:
\begin{itemize}[leftmargin=*]
    \item We provide a dataset homogenization protocol that enables Out-of-Distribution (OoD) testing of prediction algorithms across different large-scale motion datasets.
    \item We study the OoD robustness of two SotA models and explore the effect of their augmentation strategies on In-Distribution (ID) and OoD test results.
    \item We propose an efficient multi-modal predictor baseline
with competitive ID performance and superior OoD robustness by representing trajectories and map features parametrically as polynomials.
\end{itemize}

\noindent Our paper is organized as follows: We first review the recent trajectory prediction models and their design decisions with respect to data representation and augmentation strategy.  Out of the body of related work, we select two benchmark models, Forecast-MAE (FMAE) \cite{cheng_forecast_2023} and QCNet \cite{zhou_query_2023}, and detail how dataset and competition characteristics have shaped their model designs. Next, we propose our dataset homogenization protocol for OoD testing. In contrast to the benchmark design, we introduce our approach by incorporating constrained parametric representations - polynomials for both inputs and outputs. We then report ID  and OoD testing results for all three models plus a number of variants. Our results highlight the impact of different augmentation strategies in ID  and OoD testing and the increase in OoD robustness due to our choice of polynomial representations. Finally, we argue for adopting OoD testing as a performance measure of equal value to ID testing.

We open-source our code online~\cite{our_code_base}. The camera-ready version of this paper will be published in the proceedings of IROS 2024.

\section{Related Work and Benchmark Models}
\subsection{Data Representation}
\label{sec: model data representation}

The formats of individual datasets and competition rules greatly influence the design decisions of the models proposed in the literature. Models based on deep learning typically opt for a direct ingestion of the data as given.
Recent studies employ \emph{sequence-based representation}, e.g., sequences of data points, as the model's input and output \cite{cheng_forecast_2023, zhou_query_2023, liang_learning_2020, varadarajan_multipath++_2022}. This representation aligns the format of measurements in datasets and efficiently captures various information, e.g., agent trajectories and road geometries. The downside of this representation is the high redundancy and variance. The presence of measurement noise and outliers in datasets may lead to physically infeasible predictions. Additionally, the computational requirement of this approach scales with the length and temporal/spatial resolution of trajectories and road geometries.

Some previous works explored the possibility of using \emph{polynomial representations} for predictions \cite{buhet_plop_2020, su_temporally_2021}. Su et al.\ \cite{ su_temporally_2021} highlight the temporal continuity, i.e., the ability to provide arbitrary temporal resolution, of this representation. Reichardt \cite{reichardt_trajectories_2022} argues for using polynomial representations to integrate trajectory tracking and prediction into a filtering problem. Polynomial representations restrict the kind of trajectories that can be represented and introduce bias into prediction systems. This limited flexibility is generally associated with greater computational efficiency, smaller model capacity and hence better generalization. However, representing the \emph{inputs} with polynomials has not been extensively researched and deployed in recent works. In a recent study, we showed that polynomials of moderate degree can represent real-world trajectories with a high degree of accuracy \cite{yao_empirical_2023}. This result motivates the use of polynomial representations in our architecture.

\subsection{Data Augmentation}
\label{sec: model data augmentation}
Competitions typically designate one or more agents in a scenario as \emph{focal} agent and only score predictions for focal agents. However, training the model exclusively with the focal agent's behavior fails to exploit all available data. To address this, predicting the future motion of non-focal agents is a typical augmentation strategy for training. As there are many more non-focal agents than focal agents, another important design decision relates to how authors interpret the importance of focal vs.\ non-focal agent data. 

We select two open-sourced and thoroughly documented SotA models: Forecast-MAE \cite{cheng_forecast_2023} (FMAE) and QCNet \cite{zhou_query_2023}, with nearly 1900k and 7600k parameters, respectively. As summarized in Table \ref{tab: model difference in non-focal}, both models employ sequence-based representation but exhibit different strategies in dealing with non-focal agents:

\begin{itemize}[leftmargin=*]
    \item[] \textbf{Heterogeneous Augmentation}: FMAE follows the prediction competition protocol and prioritizes focal agent prediction. Thus, agent history and map information are computed within the \emph{focal agent} coordinate frame. Compared to the \emph{multi-modal} prediction of the focal agent, FMAE only outputs \emph{uni-modal} prediction for non-focal agents. The prediction error of the focal agent also weighs higher in loss function than non-focal agents.
    \item[] \textbf{Homogeneous Augmentation}: QCNet does not focus on the selected focal agent and proposes a more generalized approach. It encodes the information of agents and map elements in each agent's \emph{individual} coordinate frame. It outputs \emph{multi-modal} predictions for focal and non-focal agents alike, ensuring a consistent prediction task for all agents. The loss of focal agent prediction shares the same weight as non-focal agents.
\end{itemize}

\begin{figure}[thb]
\centering
\includegraphics[width=3.2in]{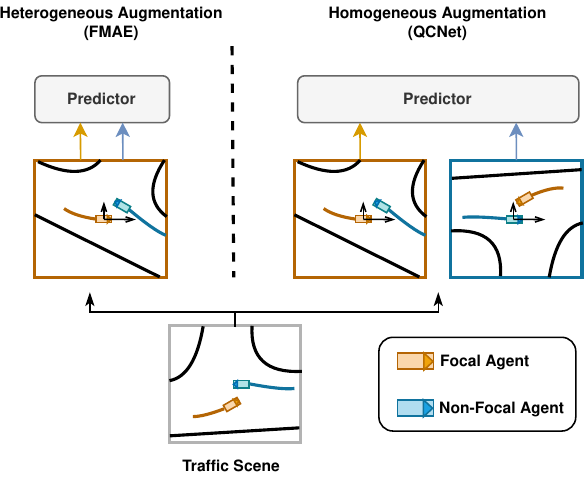}
\caption{The two augmentation strategies for non-focal agent data employed in benchmark models. \textbf{Left}: FMAE employs heterogeneous augmentation, representing information in the focal agent's coordinate frame and only making uni-modal predictions for non-focal agents. \textbf{Right}: QCNet employs homogeneous augmentation, encoding information in each agent's individual coordinate frame and making multi-modal predictions for both focal and non-focal agents alike.}
\label{fig:augmentation_strategy}
\vspace{-0.5em}
\end{figure}

\noindent Figure \ref{fig:augmentation_strategy} illustrates the augmentation strategies of our benchmark models FMAE and QCNet.

To study the influence of these augmentation approaches independently of data representation, we add a third augmentation strategy for comparison: \textbf{no augmentation}. This can simply be achieved by removing the loss of non-focal agents in each model, thus limiting the model to learn from focal agent behavior only. We denote the benchmark models without augmentation as "FMAE-noAug" and "QCNet-noAug".

For our own model, we will implement all three modes of augmentation for comparison.

\begin{table}[!tbh]
\vspace{-0.0em}
\caption{Model variants under study based on augmentation strategies and data representations.}
\centering
\begin{tabularx}{3.2in}{l >{\centering\arraybackslash}X |>{\centering\arraybackslash}X }
\Xhline{3\arrayrulewidth}
augmentation & \multicolumn{2}{!{\vrule width 1pt}c}{input and output representation} \\
\cline{2-3}
strategy&\multicolumn{1}{!{\vrule width 1pt}c|}{sequence-based} &  polynomial-based (ours) \\
\Xhline{3\arrayrulewidth}
heterogeneous & \multicolumn{1}{!{\vrule width 1pt}c|}{\multirow{1}{*}{FMAE \cite{cheng_forecast_2023}}}&\multirow{1}{*}{EP-F}\\
\cline{1-3}
homogeneous & \multicolumn{1}{!{\vrule width 1pt}c|}{\multirow{1}{*}{QCNet \cite{zhou_query_2023}}} & \multirow{1}{*}{EP-Q} \\
\cline{1-3}
\multirow{2}{*}{w/o augmentation} & \multicolumn{1}{!{\vrule width 1pt}c|}{FMAE-noAug} & \multirow{2}{*}{EP-noAug} \\
&\multicolumn{1}{!{\vrule width 1pt}c|}{QCNet-noAug}& \\
\Xhline{3\arrayrulewidth}
\end{tabularx}
\label{tab: model difference in non-focal}
\vspace{-1em}
\end{table}

\section{OoD testing and Dataset Homogenization}
The irrelevant data creation processes and platforms across motion datasets present us with the opportunity to perform OoD testing on truly independent data samples. However, this also comes with the challenges of working around inconsistencies in data formats and prediction tasks between datasets.

\begin{table*}[!th]
\vspace{5pt}
\caption{Datasets comparison and homogenization protocol for OoD testing.}
\centering
\begin{tabularx}{7.0in}{l l >{\centering\arraybackslash}X >{\centering\arraybackslash}X >{\centering\arraybackslash}X}
\Xhline{3\arrayrulewidth}
\multicolumn{2}{c}{\multirow{2}{*}{} } & \multirow{2}{*}{Argoverse 2 (A2) Dataset~\cite{wilson_argoverse2_2021}} & \multirow{2}{*}{Waymo Open Dataset (WO)~\cite{ettinger_waymo_2021}} & Homogenized Dataset\\
&&&& train \:\vline\: test\\
\Xhline{3\arrayrulewidth}
\multicolumn{2}{l}{\#samples (train / val / test)}& $\num{199908}$ / $\num{24988}$ / $\num{24984}$ & $\num{487002}$ / $\num{44097}$ / $\num{44920}$ & - \\
\hline
\multicolumn{2}{l}{\multirow{1}{*}{\#cities}}& 6 & 6 & \multirow{1}{*}{-} \\
\hline
\multicolumn{2}{l}{scenario length}& 11s & 9.1s & - \\
\hline
\multicolumn{2}{l}{sampling rate}& 10 Hz & 10 Hz & -\\
\Xhline{3\arrayrulewidth}
\multicolumn{2}{l}{history length}& 5s & 1.1s & 5s \\
\hline
\multicolumn{2}{l}{\multirow{1}{*}{prediction horizon}} & \multirow{1}{*}{6s}  & \multirow{1}{*}{8s } & 6s \:\vline\:4.1s \\
\hline
\multirow{3}{*}{\thead{map\\information}}  & lane center & yes & \multirow{1}{*}{yes} & \multirow{1}{*}{yes} \\
\cline{2-5}
& lane boundary & yes & not for junction lanes& no\\
\cline{2-5}
& junction lanes labeled & yes& no& no \\
\hline
\multirow{3}{*}{\thead{prediction\\target}}  & \# focal agents & 1 & up to 8 & 1\\
\cline{2-5}
& ego included & no & yes & no\\
\cline{2-5}
& fully observed & yes & not guaranteed & yes \\
\Xhline{3\arrayrulewidth}
\end{tabularx}
\label{tab: dataset difference}
\vspace{-1em}
\end{table*}

We propose our OoD testing protocol by training and testing models on two different large-scale datasets: A2 and WO. We summarise the characteristics of both datasets in Table \ref{tab: dataset difference}. Algorithmic improvements in generalization ability are more easily identified when training on the smaller of two datasets and testing on the larger ones. Consequently, we train on the training set from A2, comprising $\num{199908}$ samples. As the test set of the A2 and WO competitions are not accessible, they cannot be included in our homogenization protocol. Thus, for ID testing, we settle for the validation split of A2, and for OoD testing use the validation split of WO. The difference between ID and OoD results demonstrates a model's robustness against OoD test cases.




There are multiple notable distinctions between A2 and WO. One obvious distinction is the different scenario lengths: a scenario in A2 is $11$ seconds long and consists of  $5$ seconds of observation history and $6$ seconds of future to predict. WO only provides $1.1$ seconds of observation history but requires an $8$-second prediction horizon. Moreover, A2 designates and scores only one \emph{focal agent} per scenario, whereas there are up to 8 such \emph{focal agents} in WO. The focal agent in A2 is never the ego vehicle and remains fully observed throughout the prediction horizon, which is not guaranteed in WO.

To facilitate the OoD testing, we homogenize the datasets and adopt the following settings for both datasets and summarize them in Table \ref{tab: dataset difference}:
\begin{itemize}[leftmargin=*]
    \item History Length: We set the history length to 5 seconds (50 steps) as in A2.
    \item Prediction Horizon:
    We maintain the 6-second prediction horizon for training but only evaluated the first 4.1s of prediction due to the limited recording length in WO.
    \item Map Information: We exclude boundary information and the label of junction lanes due to the information's absence in WO. Only lane segments and crosswalks are considered map elements in homogenized data due to their availability in both datasets.
    \item Focal Agent: We take the same focal agent as labeled in A2. From WO, only the first fully observed, non-ego agent in the list of focal agents is chosen. As the list of focal agents is unordered, this corresponds to randomly sampling a single fully observed focal agent.
\end{itemize}

\noindent Fulfilling the focal agent's selection criterion above, we have $\num{24988}$ and $\num{42465}$ valid samples from A2 and WO validation sets for ID  and OoD testing, respectively.

\section{Proposed Model and Data Representation}
The key innovation of our model is its choice of representation for map elements and trajectories on both input and output sides of the model.
Here, we propose to use a parametric representation in terms of polynomials. This constrains the model's expressiveness compared to sequence-based data representations. Such a parametric representation is easily pre-computed for map elements and algorithms for online tracking and filtering agent observations into trajectories already exist \cite{reichardt_trajectories_2022}. 

Our prior work on trajectory representation \cite{yao_empirical_2023} has shown that polynomials are an accurate representation of agent trajectories and should not present a limiting factor to the achievable prediction performance. Independent research suggests that sequence-based data representations may carry high-frequency information, which can result in over-fitting \cite{yin_fourier_2019}.

With the polynomial inputs and outputs as our key innovation, we present our approach with great model efficiency and generalization: \emph{Everything Polynomial} (EP). We now elaborate our approach to represent data with polynomials. 

\subsection{Inputs as Polynomials}
We employ Bernstein polynomials to represent the agent history and map geometry. The parameters of Bernstein polynomials have spatial semantics as \emph{control points}.

\emph{Agent History}: Based on the Akaike Information Criterion (AIC) \cite{akaike_AIC_1973} presented in \cite{yao_empirical_2023}, the 5-second history trajectories of vehicles, cyclists, and pedestrians in A2 and WO are represented optimally with 5-degree polynomials. We also use the 5-degree polynomial for the ego vehicle.
We track the control points of agent history with the method proposed in \cite{reichardt_trajectories_2022} and incorporate the observation noise models proposed in \cite{yao_empirical_2023}. 

\emph{Map}: Map elements, such as lane segments
and crosswalks, are represented with 3-degree polynomials - a degree consistent with OpenDRIVE \cite{opendrive} standards. We fit the sample points of map elements via the total-least-squares method by Borges-Pastva \cite{borges_2002_total}. Lane segments in WO are longer than those in A2, posing a challenge for fitting them with 3-degree polynomials. Therefore, we iteratively split the lane segments by half until the fit error is under $0.1$m.

\begin{figure}[bth]
\centering
\includegraphics[width=3.2in]{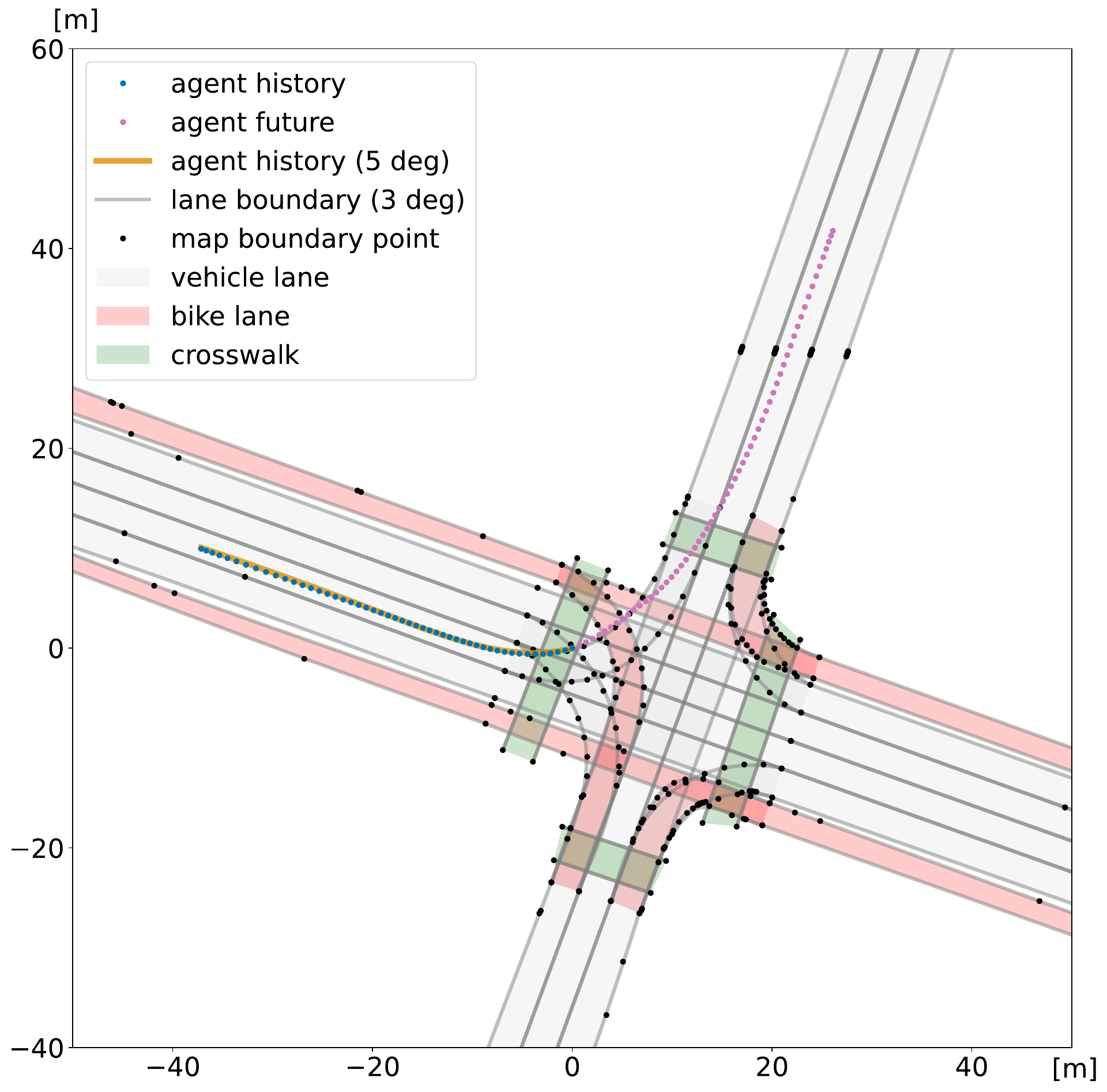}
\caption{One traffic scenario in A2 represented with polynomials.}
\label{fig:A2_fitted}
\vspace{-1.0em}
\end{figure}

We visualized one traffic scenario in A2 with polynomial representations in Figure \ref{fig:A2_fitted}. The polynomial representation only requires $40.8 \%$ and $8.7 \%$ of data space compared to the sample points provided in A2 and WO, respectively.

\begin{table*}[t]
\vspace{+5pt}
\caption{Result of In-Distribution testing (6-second prediction horizon) in competition setting on Argoverse 2 test set. For all the metrics, lower the better. Inference time is tested on one Tesla T4 GPU with a scenario of 50 agents and 150 map elements.}
\centering
    \begin{tabularx}{\textwidth}{l *{4}{>{\centering\arraybackslash}X} | cc}
    \toprule
    \multirow{2}{*}{\shortstack{model}} & minADE$_{1}$ & minFDE$_{1}$ & minADE$_{6}$ & minFDE$_{6}$ & \# model & inference time\\
        & [m]&[m]& [m] & [m] & params [k]& [ms]\\
        \midrule
        \multirow{2}{*}{QCNet \cite{zhou_query_2023}} &1.702 & 4.309 & 0.643& 1.244 &7600& 120.89\\
        &(100.0\%)&(100.0\%)&(100.0\%)&(100.0\%)&(100.0\%)&(100.0\%)\\
        \cmidrule(lr){2-5} \cmidrule(lr){6-7}
        Forecast-MAE \cite{cheng_forecast_2023} &1.741 & 4.355& 0.709 & 1.392 & 1900 &12.63\\
        w pre-train &(102.2\%)&(101.1\%)&(110.3\%)&(111.9\%)&(25.0\%)&(10.4\%)\\
        \cmidrule(lr){2-5} \cmidrule(lr){6-7}
        \multirow{2}{*}{EP-Q (ours)} &2.134 & 5.415 & 0.841 & 1.683 & 334 & 5.72\\
        &(125.4\%)&(125.7\%)&(131.0\%)&(135.3\%)&(4.4\%)&(4.7\%)\\
        \cmidrule(lr){2-5} \cmidrule(lr){6-7}
        \multirow{2}{*}{EP-F (ours)} & 1.887 & 4.567  & 0.801 & 1.526 & \multirow{4}{*}{\thead{345 \\(4.5\%)}}&\multirow{4}{*}{\thead{4.66\\ (3.9\%)}}\\ 
        &(110.9\%) & (106.0\%) & (124.6\%) & (122.7\%) &&\\
        \cmidrule(lr){2-5}
        \multirow{2}{*}{EP-F w flip (ours)} & 1.846 & 4.456 & 
        0.786& 1.485& &\\
        & (108.5\%) & (103.4\%) & (122.2\%) & (119.4\%)&&\\
        \bottomrule
    \end{tabularx}
    \label{tab: in-distribution result}
    \vspace{-5pt}
\end{table*}

\subsection{Outputs as Polynomials}
The 6-second predicted trajectories are formulated as 6-degree polynomials, one degree higher than indicated by AIC in \cite{yao_empirical_2023}. Instead of predicting the polynomial parameters of future trajectories, our model predicts several future kinematic vehicle states. By fusing the predicted future states with current tracked states, we recalculate the polynomial parameters and reconstruct the future trajectory. Details are presented in Appendix \ref{sec: ep implementation}. 

\subsection{Model}
The pipeline of EP is illustrated in Figure \ref{fig:pipeline}.
EP employs the popular encoder and decoder architecture and its implementation details are presented in Appendix \ref{sec: ep implementation}. 

Following the different augmentation strategies by our benchmark models, we implemented three EP variants: (1) EP-F employs the heterogeneous augmentation of FMAE. (2) EP-Q employs the homogeneous augmentation of QCNet. (3) EP-noAug has no augmentation.  Table \ref{tab: model difference in non-focal} summarizes the design choices of the different models under study.

Due to the compact polynomial representations for inputs and outputs, EP variants only employ around 345k model parameters, which is only $4.5\%$ the size of QCNet and $18.2\%$ the size of FMAE as listed in Table \ref{tab: in-distribution result}.

Since we focus on OoD robustness, scaling up our model for more model capacity and potentially better ID performance is not prioritized in this work.

\section{Experimental Results}
\subsection{Experiment Setup on Homogenized Data}
On the homogenized A2 training data, all models are trained from scratch with hyperparameters reported for A2 competition performance. Due to our dataset homogenization settings, we have to adjust the data pre-processing and corresponding encoding layers in QCNet, e.g., removing the layers for encoding the label of junction lanes. FMAE's architecture is not affected by our data homogenization. Training settings of EP variants are summarized in Appendix \ref{sec: EP Experiment Settings}.

\subsection{Metrics}
\label{sec: metrics}
For ID testing, we use the official benchmark metrics, including minimum Average Displacement Error ($minADE_{K}$) and minimum Final Displacement Error
($minFDE_{K}$), for evaluation. The metric $minADE_{K}$ calculates the Euclidean distance in meters between the ground-truth trajectory and the best of $K$ predicted trajectories as an average of all future time steps. Conversely, $minFDE_{K}$ focuses solely on the prediction error at the final time step, emphasizing long-term performance. 

For OoD testing, we also propose $\Delta minADE_{K}$ and $\Delta minFDE_{K}$ as the difference of displacement error between ID  and OoD testing to measure model robustness. In accordance with standard practice, $K$ is selected as 1 and 6.

\subsection{ID Results with Competition Settings}
Like the other studies, we first present the best ID test results of all models according to the A2 prediction competition protocol in Table \ref{tab: in-distribution result}. Benchmark results are from the original authors. For EP, we also include the results when augmenting the data by symmetrizing around the x-axis, i.e. flipping all left turns into right turns and vice versa. 

Though the performance in ID testing on the A2 test set is not the focus of our work, our EP still achieves near state-of-the-art performance with a significantly smaller model size. For instance, with ``flip" data augmentation, the $minFDE_{1}$ of EP-F is only $3.4\%$ higher compared to QCNet. Though EP exhibits a considerable gap compared to QCNet in terms of multi-modal prediction (K=6), it exhibits only less than $10$cm additional prediction error in $minADE_{6}$ and $minFDE_{6}$ compared to FMAE with pre-train. EP has a faster inference time, requiring only $3.9\%$ of QCNet's and $36.9\%$ of FMAE's inference time, which is essential in real-time applications. Another notable point is the better ID performance of EP-F compared to EP-Q. This indicates that heterogeneous augmentation shows more effectiveness than
homogeneous augmentation in ID testing when only a single focal agent is scored. 

\begin{figure}[!bth]
\centering
\includegraphics[width=3.4in]{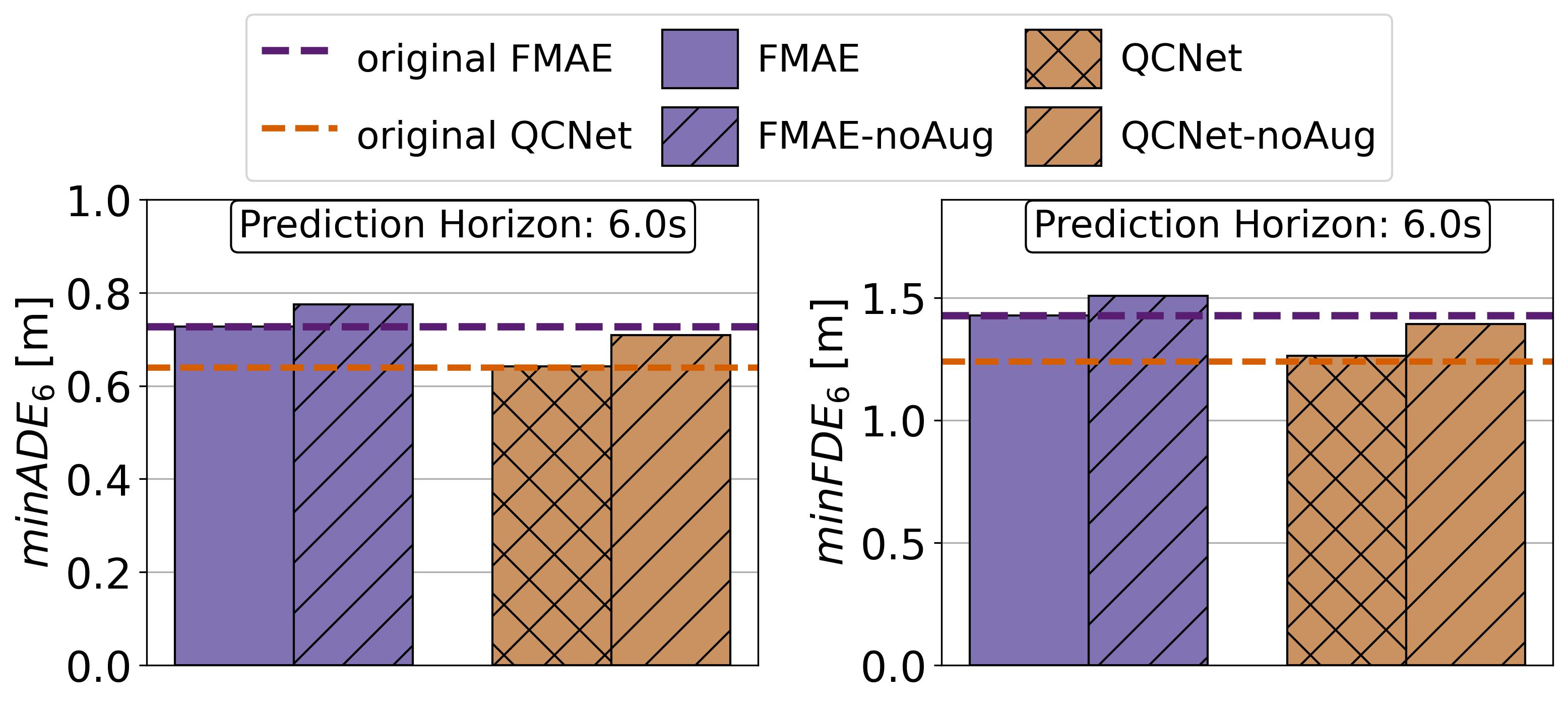}
\caption{ID performance of benchmarks on A2. \textbf{Dashed Lines}: Results reported by original authors with competition settings. \textbf{Bars}: Results for retrained benchmark models on homogenized datasets.}
\label{fig:ID_results}
\vspace{-0.5em}
\end{figure}

Next, we examine the impact of dataset homogenization and augmentation. In Figure \ref{fig:ID_results}, we can see that the alterations due to dataset homogenization, e.g., excluding lane boundaries, have only little effect on model performance. Removing the data augmentation, however, has a small but noticeable negative effect on ID performance for both benchmark models. These results also show that our retraining works as expected.

\label{sec: ID_results}

\subsection{OoD Testing Results}
\label{sec: OoD_results}
The key question now is whether the results achieved in ID testing on A2 will translate to OoD testing on WO. Figure \ref{fig:OoD_results} summarizes the results. As
discussed, the relative and absolute increase in the prediction metric will serve as our measure of robustness and generalization ability. We will present OoD results from three perspectives: (1) augmentation strategy, (2) data representation, and (3) contrast results between ID  and OoD testing.

\subsubsection{Augmentation Strategy}
We observe that without augmentation, i.e., excluding the non-focal agents in the loss function, all models generalize poorly on all metrics, but our model has the smallest relative and absolute increase in error in all cases. 

Though heterogeneous augmentation does marginally reduce FMAE's prediction error for both ID  and OoD testing, the relative and absolute increase in prediction error between ID and OoD samples is practically the same as in the non-augmented case, e.g., $+0.279$m $(61.5\%)$ vs. $+0.295$m $(69.7\%)$ in $\Delta minADE_{6}$.

\begin{figure}[!bth]
\centering
\includegraphics[width=3.2in]{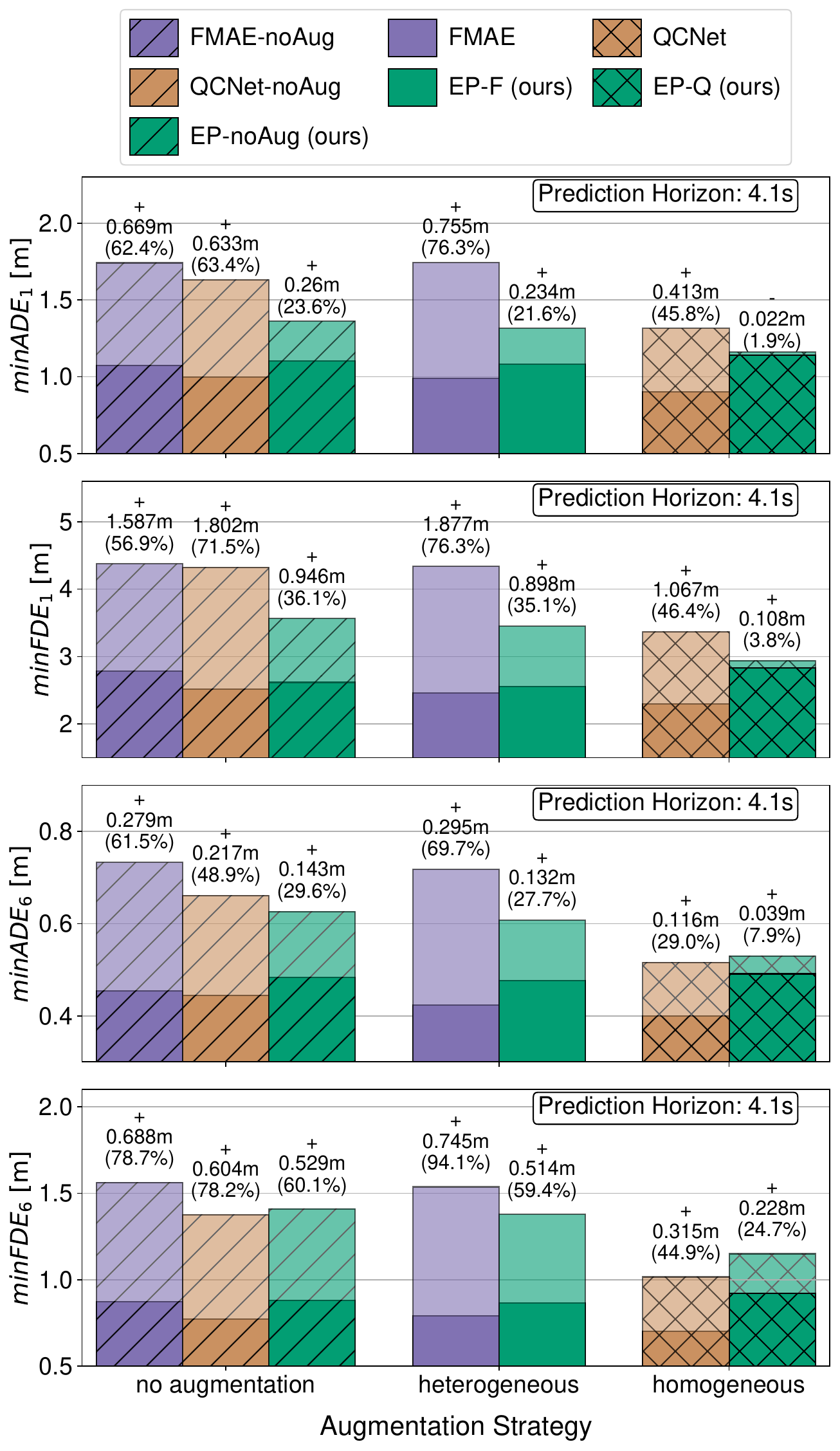}
\caption{The OoD testing results of FMAE, QCNet, EP and their variants, as reported in Table \ref{tab: ID and OoD result homo} in Appendix \ref{sec: ID Results homo}. We indicate the \emph{absolute} and \emph{relative} difference in displacement error between ID  and OoD results. \textbf{Solid}: The ID results by training on the homogenized A2 training set and testing on the homogenized A2 validation set. \textbf{Transparent}: The increased displacement error in OoD testing by training on the homogenized A2 training set and testing on the homogenized WO validation set.}
\label{fig:OoD_results}
\vspace{-1.0em}
\end{figure}

On the other hand, the robustness of QCNet significantly benefits from the homogeneous augmentation with reduced $\Delta minFDE_{1}$ from $+1.802$m $(71.5\%)$ to $+1.067$m $(46.4\%)$. Compared to EP-noAug, EP-F and EP-Q replicate the behavior observed from FMAE and QCNet, e.g., with only a marginal impact from heterogeneous augmentation and a significant impact from homogeneous augmentation.


\subsubsection{Data Representation}
Our EP variants exhibit improved robustness compared to benchmarks with sequence-based data independent of data augmentation strategy. For instance, EP-F outperforms FMAE by demonstrating lower prediction errors and a lower increase in prediction errors in OoD testing. Similarly, compared to QCNet, EP-Q also demonstrates improved robustness, even achieving a reduction in prediction error of $-0.022$m $(-1.9\%)$ in $\Delta minADE_1$.

\subsubsection{ID vs OoD results}
We observe multiple reversals in performance between ID and OoD testing results: (1) Despite FMAE demonstrating comparable ID performance to QCNet, it exhibits significantly lower robustness than QCNet in OoD testing. (2) While FMAE and QCNet outperform EP on ID samples, EP shows lower prediction error than FMAE and QCNet across multiple metrics in OoD testing. (3) There is a reversal in performance between EP-F and EP-Q in both ID and OoD testing.

We also examine the impact of model size on OoD results and scale the baselines to a comparable size to EP. Details and results are reported in Appendix \ref{sec: scaled baseline}. Compared to the scaled baselines, our model still demonstrates improved robustness and even outperforms the scaled QCNet with lower prediction errors in $minADE_6$ and $minFDE_6$ in OoD testing. 

\begin{table}[!tbh]
\caption{ID and OoD testing results (4.1-second prediction horizon) of EP-F and EP-F (w flip).}
\vspace{-0.0em}
\centering
\begin{tabularx}{3.4in}{l >{\centering\arraybackslash}X >{\centering\arraybackslash}X }
\Xhline{3\arrayrulewidth}
\multirow{2}{*}{model}&  \multicolumn{2}{c}{$minADE_{1} | minFDE_{1}$ [m]} \\
& ID & OoD \\
\Xhline{3\arrayrulewidth}
EP-F& 1.08 \vline \: 2.56&  1.32 \vline  \: 3.45\\
\cline{1-3}
EP-F (w flip) & 1.06 \vline \: 2.50 & 1.35 \vline \: 3.53 \\
\Xhline{3\arrayrulewidth}
\end{tabularx}
\label{tab: data flip impact}
\vspace{-1em}
\end{table}

Moreover, we also note the contrasting impact of "data flipping" in ID and OoD testing, as summarized in Table \ref{tab: data flip impact}. This indicates that improper data augmentation can in fact impair model generalization.

The differences between ID and OoD results underscore the importance of OoD testing for a thorough evaluation of trajectory prediction models.





\section{Conclusion}    
Recent SotA trajectory prediction models have thoroughly optimized their In-Distribution (ID) performance and present outstanding results in test sets evaluated in individual prediction competition. However, better ID performance does not automatically guarantee higher Out-of-Distribution (OoD) performance. In fact, performance differences can and do reverse in OoD settings. Our proposed OoD testing protocol enables a fresh perspective for model evaluation in trajectory prediction that we hope the community will adopt.
With our protocol, we demonstrate the robustness improvement from \emph{homogenous augmentation} and prove the benefits of \emph{polynomial representation} as employed in our EP model. With a much smaller model size and lower inference time, EP achieves near SotA ID performance and exhibits significantly improved OoD robustness compared to benchmark models. This work represents an initial step toward a trajectory prediction model that is capable of generalizing to different datasets and is robust under changes in sensor setup, scenario selection strategy or post-processing of training data.
More sophisticated models based on this concept will be developed to improve the performance for both In- and Out-of-Distribution evaluations.

\begin{samepage}
\section*{ACKNOWLEDGMENT}
Authors would like to thank Andreas Philipp for many useful discussions and acknowledge funding from the German Federal Ministry for Economic Affairs and Climate Action within the project “KI Wissen – Automotive AI powered by Knowledge”.
\end{samepage}

\bibliographystyle{IEEEtran}
\bibliography{References}

\begin{appendices}

\section{Implementation Details of Our Model}
\label{sec: ep implementation}
\subsection{Model}
\subsubsection{Input}
The control points of $A$ agents in either \emph{focal agent} or \emph{individual} coordinate are denoted as $\boldsymbol{\omega}^{a}_{n} \in \mathbb{R}^{A \times 2}$ with $n\in\{0,1,\cdots,5\}$. Following the popular vectorization approach, we compute the vector between control points as $\boldsymbol{\delta}^{a}_{n} = \boldsymbol{\omega}^{a}_{n} - \boldsymbol{\omega}^{a}_{n-1}$ for capturing the kinematic of agents. To describe the spatial relationship, we employ the last control point $\boldsymbol{\omega}^{a}_{5}$ and the normalized vector $\boldsymbol{\theta}^{a}_{5} = norm(\boldsymbol{\delta}^{a}_{5})$ as the reference position and orientation for agents. The time window of agent appearance is denoted as $\boldsymbol{TW} \in \mathbb{R}^{A \times 2}$. Therefore, the features of agents include $\boldsymbol{\Delta}^{a} = [\boldsymbol{\delta}^{a}_{1}, \boldsymbol{\delta}^{a}_{2}, \cdots, \boldsymbol{\delta}^{a}_{5}]$, $\boldsymbol{PI}^{a}=[\boldsymbol{\omega}^{a}_{5}, cos(\boldsymbol{\theta}^{a}_{5}), sin(\boldsymbol{\theta}^{a}_{5})]$ and  $\boldsymbol{TW}$. The features of $M$ map elements are computed similarly with considering the initial point as the reference, denoting as $\boldsymbol{\Delta}^{m} = [\boldsymbol{\delta}^{m}_{1}, \boldsymbol{\delta}^{m}_{2}, \boldsymbol{\delta}^{m}_{3}]$ and $\boldsymbol{PI}^{m}=[\boldsymbol{\omega}^{m}_{0}, cos(\boldsymbol{\theta}^{m}_{1}), sin(\boldsymbol{\theta}^{m}_{1})]$.

\begin{figure}[!bth]
\centering
\includegraphics[width=3.2in]{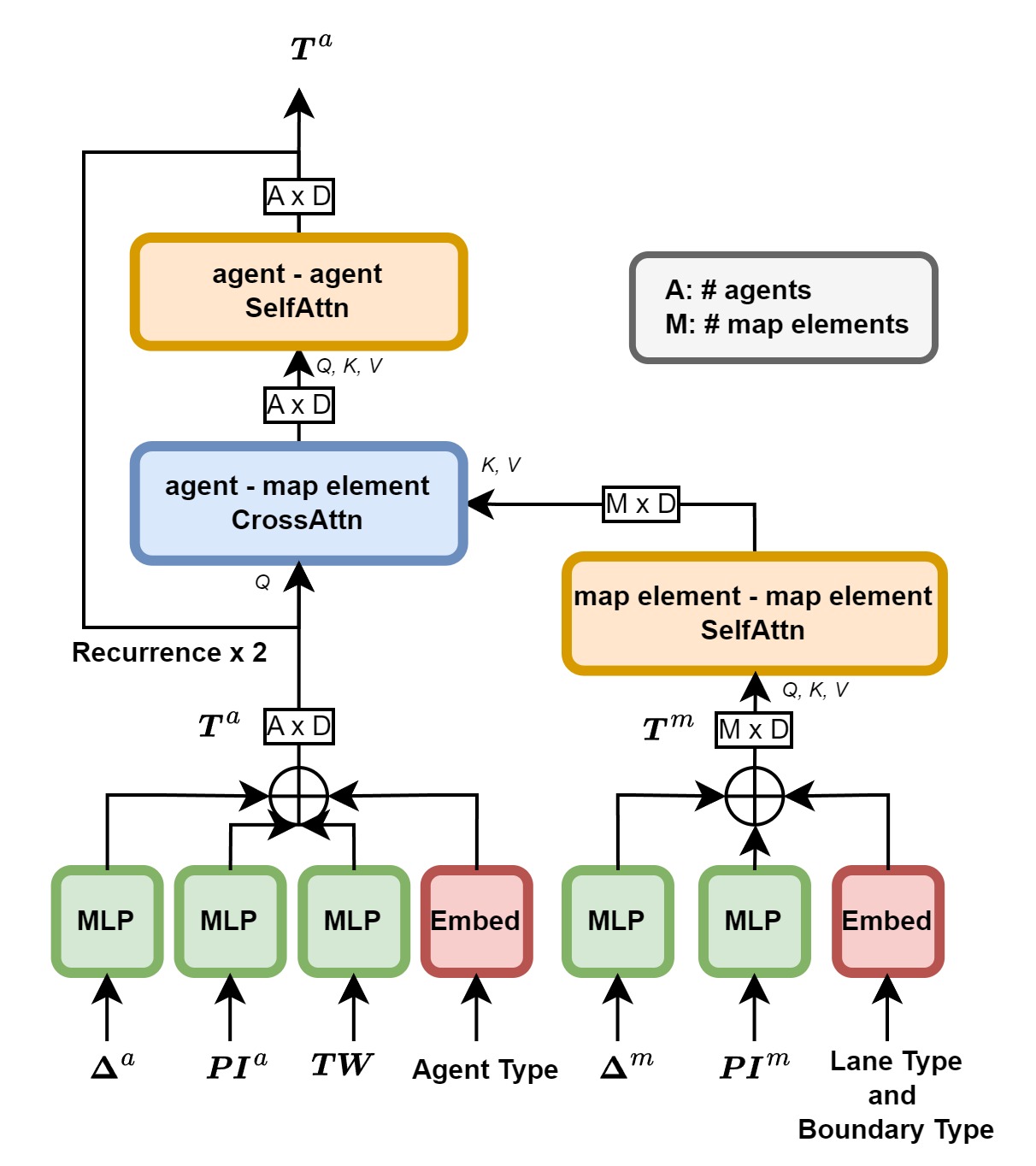}
\caption{Encoder architecture of EP variants.}
\label{fig:encoder architecture}
\vspace{-1.0em}
\end{figure}

\subsubsection{Encoder}
As visualized in Figure \ref{fig:encoder architecture}, we employ simple 3-layer MLPs to encode the features of agents and map elements. Semantic attributes, such as agent type and lane type, are encoded with individual embedding layers. Embedded features are added and fused to agent tokens $\boldsymbol{T}^{a} \in \mathbb{R}^{A \times D}$ and map element tokens $\boldsymbol{T}^{m} \in \mathbb{R}^{M \times D}$, where $D$ denotes the hidden dimension. Multiple attention blocks based on Transformer \cite{vaswani_attention_2017} and "pre-layer normalization" \cite{xiong_layer_2020} perform the \emph{map element}-\emph{map element}, \emph{agent}-\emph{map element} and \emph{agent}-\emph{agent} attentions sequentially and update $\boldsymbol{T}^{a}$. All EP variants share the same encoder architecture.

\begin{figure}[!bth]
\centering
\includegraphics[width=3.2in]{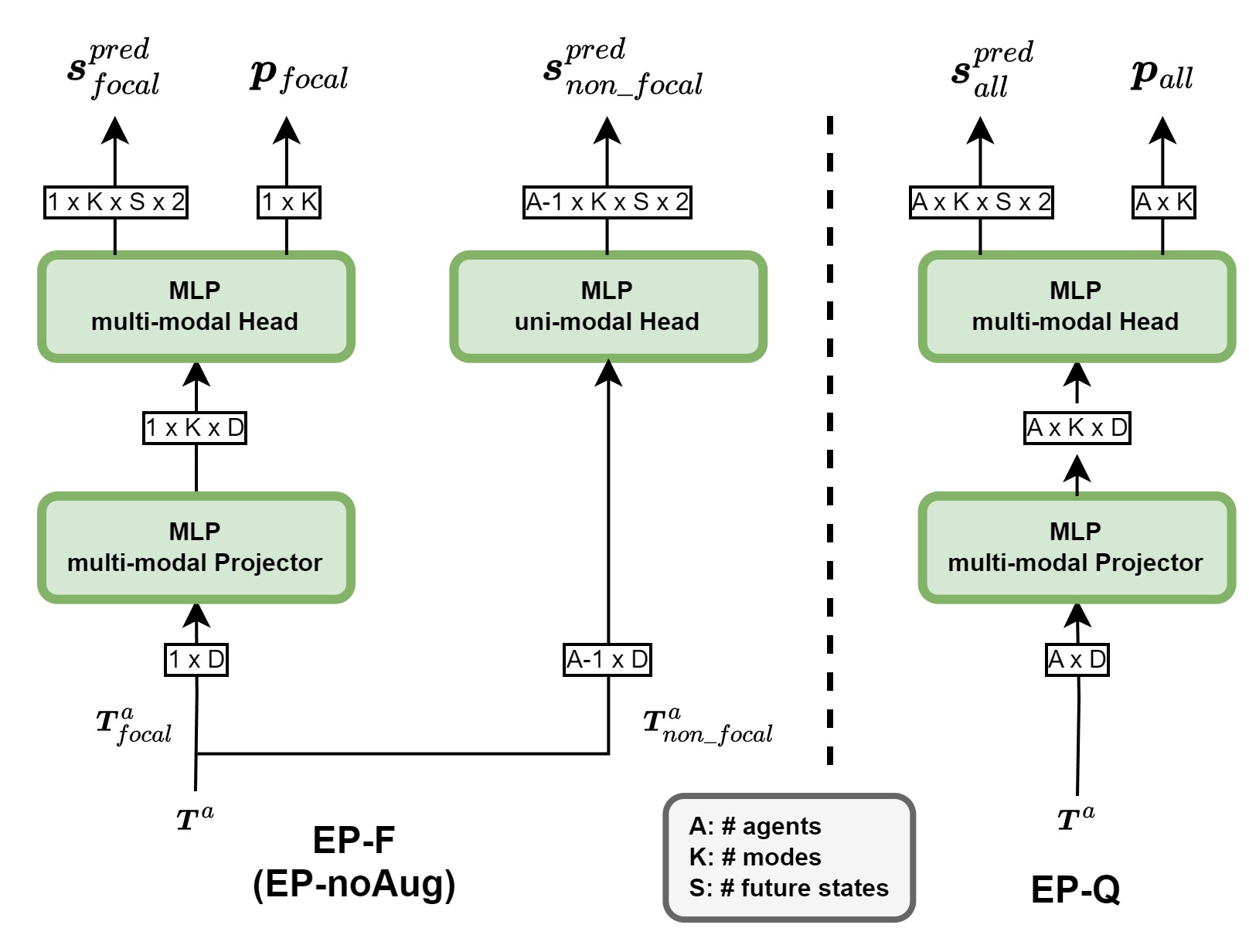}
\caption{Decoder architectures of EP-F, EP-Q and EP-noAug.}
\label{fig:decoder architecture}
\vspace{-1.0em}
\end{figure}

\subsubsection{Decoder}
In Figure \ref{fig:decoder architecture}, we visualize the two decoder architectures of EP-F (EP-noAug) and EP-Q due to different augmentation strategies. 

The embedded agent features can be decoded in two ways: (1) For multi-modal prediction, the agent token $\boldsymbol{T}^{a}$ is firstly projected to multiple modes and substantially decoded to predicted states $\boldsymbol{s}^{pred}$ and probabilities $\boldsymbol{p}$ for each mode. (2) For uni-modal prediction, the agent token $\boldsymbol{T}^{a}$ is directly decoded to the predicted states of a single mode. 

Consider the $k$-th mode of $i$-th agent, we concatenate the two-dimensional states as the vector $\boldsymbol{s}_{i, k} = [\boldsymbol{p}^{track}_{i, 0}, \boldsymbol{p}^{pred}_{i, k, 30}, \boldsymbol{v}^{pred}_{i, k, 30}, \boldsymbol{a}^{pred}_{i, k, 30}, \boldsymbol{p}^{pred}_{i, k, 60}, \boldsymbol{v}^{pred}_{i, k, 60}, {a}^{pred}_{i, k, 60}], \boldsymbol{s}_{i, k} \in \mathbb{R}^{14}$, where $\boldsymbol{p}^{track}_{i, 0}$ denotes the tracked position at current time step, $\boldsymbol{p}^{pred}_{i, k, 30}, \boldsymbol{v}^{pred}_{i, k, 30}, \boldsymbol{a}^{pred}_{i, k, 30}$ are the predicted position, velocity and acceleration at 3s (30 steps) in the future, respectively. $\boldsymbol{p}^{pred}_{i, k, 60}, \boldsymbol{v}^{pred}_{i, k, 60}, {a}^{pred}_{i, k, 60}$ are the predicted position, velocity and acceleration at 6s (60 steps) in the future, respectively.

We follow the same notation as in \cite{yao_empirical_2023} and form the monomial basis functions as a vector $\boldsymbol{\phi}(t_{m})\in \mathbb{R}^{7}$ by concatenation $\boldsymbol{\phi}(t_{m})=[\phi_0(t_{m}),\cdots,\phi_6(t_{m})]^{\top}$, where $t_{m}$ denotes the timestamp at $m$-th step in the future. The derivatives of the monomial basis function $\boldsymbol{\dot{\phi}}(t{m})$ and $\boldsymbol{\ddot{\phi}}(t_{m})$ can be acquired by incorporating the linear derivative operator $\boldsymbol{D} \in \mathbb{R}^{7\times 7}$ mentioned in \cite{reichardt_trajectories_2022}. We construct the $14\times 14$ observation matrix $\boldsymbol{H}=[\boldsymbol{\phi}(t_{0}), \boldsymbol{\phi}(t_{30}), \dot {\boldsymbol{\phi}}(t_{30}), \ddot{\boldsymbol{\phi}}(t_{30}), \boldsymbol{\phi}(t_{60}), \dot{\boldsymbol{\phi}}(t_{60}), \ddot{\boldsymbol{\phi}}(t_{60})]^{\top}\otimes\mathbf{I}_2$, where $\mathbf{I}_2$ is a $2\times2$ identity matrix and $\otimes$ denotes the Kronecker product. The trajectory parameters $\boldsymbol{\omega}_{i, k}$ of the $k$-th predicted mode are expressed as $\boldsymbol{\omega}^{pred}_{i, k}=(\boldsymbol{H}^{\top}\boldsymbol{H})^{-1}\boldsymbol{H}^{\top}\boldsymbol{s}_{i, k}$.

\subsection{Training Loss}
We employ $minADE_{6}$ as regression loss $l^{reg, multi}$ and weighted average displacement error, with the weight of predicted probability $\boldsymbol{p}$, as classification loss $l^{cls, multi}$ for multi-modal prediction. We calculate average displacement error as regression loss for uni-modal prediction $l^{reg, uni}$. Thus, the training losses of EP variants are expressed as:
\begin{equation}
\label{eqn:EP loss}
\begin{aligned}
l_{EP-F} = & l^{reg, multi}_{focal} + l^{cls, multi}_{focal} + l^{reg, uni}_{non\_focal}\\
l_{EP-Q} = & l^{reg, multi}_{all} + l^{cls, multi}_{all}\\
l_{EP-noAug} = &  l^{reg, multi}_{focal} + l^{cls, multi}_{focal}\\
\end{aligned}
\end{equation}
where sub-indices "focal," "non-focal," and "all" refer to the focal agent, non-focal agents, and all agents, respectively.

\subsection{Experiment Settings}
\label{sec: EP Experiment Settings}
We report the settings of EP variants in Table \ref{tab: setting for EP}. Note that the batch size and number of epochs are designed to ensure the same training iterations for all EP variants. 
\begin{table}[!tbh]
\caption{Settings for EP Variants}
\vspace{-0.0em}
\centering
\begin{tabularx}{3.2in}{c  >{\centering\arraybackslash}X |>{\centering\arraybackslash}X }
\Xhline{3\arrayrulewidth}
model& EP-F \& EP-noAug & EP-Q \\
\Xhline{3\arrayrulewidth}
hidden dimension $D$ &  \multicolumn{2}{c}{64}\\
\hline
optimizer &  \multicolumn{2}{c}{Adam}\\
\hline
learning rate & 1e-3 & 5e-4\\
\hline
learning rate schedule &  \multicolumn{2}{c}{cosine}\\
\hline
batch size &  64 (128 w flip)& 32\\
\hline
training epochs &  128 & 64\\
\hline
warmup iterations &  \multicolumn{2}{c}{6e4} \\
\Xhline{3\arrayrulewidth}
\vspace{-1em}
\end{tabularx}
\label{tab: setting for EP}
\end{table}
\section{Scaling of Baselines}
\label{sec: scaled baseline}
\subsection{Scaling Details of Baselines}
We scale FMAE and QCNet to a comparable size to EP, with 313k and 413k model parameters, respectively. Scaling details are reported in Tables \ref{tab: baseline scaling FMAE} and \ref{tab: baseline scaling QCNet}. Other hyper-parameters remain the same for training the scaled baselines.

\begin{table}[!h]
\caption{Scaling Details of FMAE}
\centering
\begin{tabularx}{3.2in}{c  >{\centering\arraybackslash}X }
\Xhline{3\arrayrulewidth}
parameter & modification  \\
\Xhline{3\arrayrulewidth}
embed dim & $128 \rightarrow 48$\\
\hline
num heads & $8 \rightarrow 4$  \\
\hline
embed dim & $512 \rightarrow 180$  \\
in \emph{lane\_embedding} layers & $256 \rightarrow 90$  \\
\Xhline{3\arrayrulewidth}
\end{tabularx}
\label{tab: baseline scaling FMAE}
\end{table}

\begin{table}[!h]
\caption{Scaling Details of QCNet}
\centering
\begin{tabularx}{3.2in}{c  >{\centering\arraybackslash}X }
\Xhline{3\arrayrulewidth}
parameter & modification  \\
\Xhline{3\arrayrulewidth}
hidden dim & $128 \rightarrow 32$\\
\hline
head dim & $16 \rightarrow 8$\\
\hline
num heads & $4 \rightarrow 3$  \\
\hline
num agent layers & $2 \rightarrow 1$  \\
\hline
num dec layers & $2 \rightarrow 1$  \\
\Xhline{3\arrayrulewidth}
\end{tabularx}
\label{tab: baseline scaling QCNet}
\end{table}

\newpage
\subsection{OoD Testing Results with Scaled Baselines}

\begin{figure}[!th]
\centering
\includegraphics[width=3.2in]{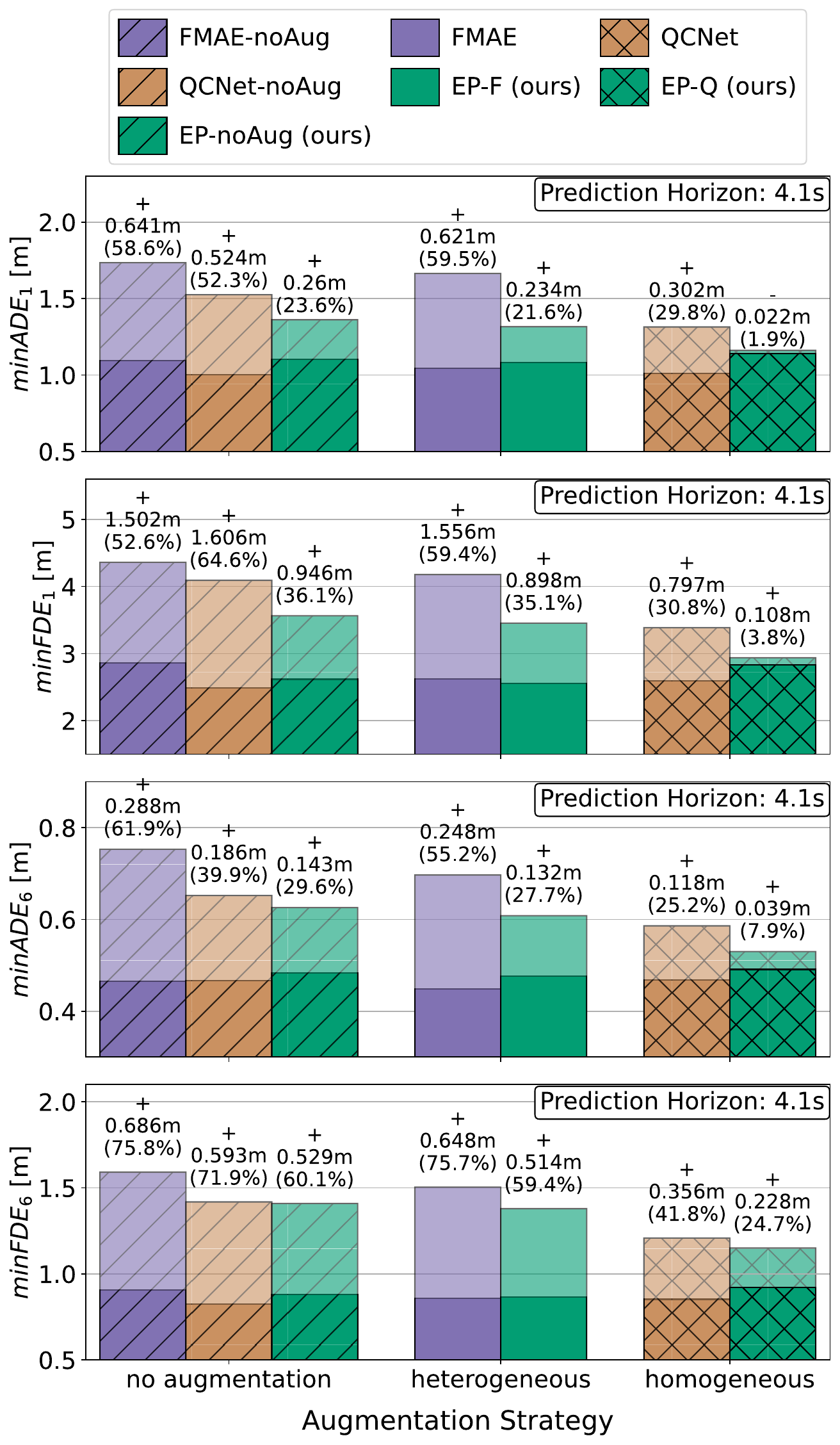}
\caption{The OoD testing results of \textbf{scaled} FMAE (313k), \textbf{scaled} QCNet (413k), EP and their variants. We indicate the \emph{absolute} and \emph{relative} difference in displacement error between ID  and OoD results. \textbf{Solid}: The ID results by training on the homogenized A2 training set and testing on the homogenized A2 validation set. \textbf{Transparent}: The increased displacement error in OoD testing by training on the homogenized A2 training set and testing on the homogenized WO validation set.}
\label{fig:OoD_results}
\end{figure}

\section{ID and OoD Results}
\label{sec: ID Results homo}

We report both ID and OoD results for all models tested on homogenenized datasets in Table \ref{tab: ID and OoD result homo}. 
\newpage
\noindent
\begin{table*}[thb]
\centering
\caption{Results of In-Distribution (ID) and Out-of-Distribution (OoD) testing of QCNet, FMAE, EP and their variants evaluated with a 4.1-second prediction horizon. \textbf{Upper Section}: ID result of models trained on the homogenized A2 training set and tested on the homogenized A2 validation set. \textbf{Lower Section}: OoD result of models trained on the homogenized A2 training set and tested on the homogenized WO validation set.}
    \begin{tabularx}{\textwidth}{l *{5}{>{\centering\arraybackslash}X}}
    \toprule
    \multirow{2}{*}{\shortstack{Testing}} & \multirow{2}{*}{\shortstack{model}} & minADE$_{1}$ & minFDE$_{1}$ & minADE$_{6}$ & minFDE$_{6}$ \\
        && [m]&[m]& [m] & [m] \\
        \midrule
        &QCNet-noAug \cite{zhou_query_2023} &0.998 & 2.519 & 0.444& 0.772 \\
        \cmidrule(lr){2-6}
        &QCNet \cite{zhou_query_2023} &0.902 & 2.301 & 0.400& 0.701 \\
        \hhline{~=====}
        &FMAE-noAug \cite{cheng_forecast_2023} &1.072 & 2.787& 0.454 & 0.874 \\
       \cmidrule(lr){2-6}
        A2 (ID)&FMAE \cite{cheng_forecast_2023} &0.989 & 2.459& 0.423 & 0.792 \\
        \hhline{~=====}
        &EP-noAug (ours) &1.101 & 2.617 & 0.483 & 0.880 \\
       \cmidrule(lr){2-6}
        &EP-F (ours) & 1.082 & 2.555  & 0.476 & 0.865 \\ 
        \cmidrule(lr){2-6}
        &EP-Q (ours) &1.161 & 2.830 & 0.491 & 0.922 \\
        \bottomrule
        &QCNet-noAug \cite{zhou_query_2023} &1.631 & 4.321 & 0.661& 1.376 \\
        \cmidrule(lr){2-6}
        &QCNet \cite{zhou_query_2023} &1.315 & 3.368 & 0.516& 1.016 \\
        \hhline{~=====}
        &FMAE-noAug \cite{cheng_forecast_2023} &1.741 & 4.374& 0.733 & 1.562 \\
       \cmidrule(lr){2-6}
        WO (OoD)&FMAE \cite{cheng_forecast_2023} &1.744 & 4.336& 0.718 & 1.537 \\
        \hhline{~=====}
        &EP-noAug (ours) &1.361 & 3.563 & 0.626 & 1.409 \\
       \cmidrule(lr){2-6}
        &EP-F (ours) & 1.316 & 3.453  & 0.608 & 1.379 \\ 
        \cmidrule(lr){2-6}
        &EP-Q (ours) &1.139 & 2.938 & 0.530 & 1.150 \\
        \bottomrule
    \end{tabularx}
    \label{tab: ID and OoD result homo}
\end{table*}

\end{appendices}
\end{document}